\documentclass[11pt, a4paper]{article}
\usepackage[utf8]{inputenc}
\usepackage{amsmath}
\usepackage{geometry}
\usepackage{helvet} 
\usepackage{charter} 
\usepackage{xcolor}
\usepackage{titlesec}
\usepackage{booktabs}
\usepackage{graphicx}
\usepackage{listings}
\usepackage[export]{adjustbox}

\definecolor{techblue}{HTML}{005F73}

\newcommand*{\inlineimage}[1]{%
  \raisebox{-.3\baselineskip}{%
    \includegraphics[
      height=15em,
      valign=c,
      keepaspectratio,
    ]{#1}%
  }%
}

\newcommand*{\inlineimagemed}[1]{%
  \raisebox{-.3\baselineskip}{%
    \includegraphics[
      height=12.5em,
      valign=c,
      keepaspectratio,
    ]{#1}%
  }%
}

\newcommand*{\inlineimagesmall}[1]{%
  \raisebox{-.3\baselineskip}{%
    \includegraphics[
      height=10em,
      valign=c,
      keepaspectratio,
    ]{#1}%
  }%
}

\titleformat{\section}{\color{black}\normalfont\Large\bfseries}{\thesection}{1em}{}

\title{\textbf{A Consensus-Based Framework for Relative Preference Evaluation of Large Language Models}}
\author{Mohtashim Khan}
\date{April 2026}

\begin{document}

\maketitle

\begin{abstract}
Traditional benchmarks for LLMs primarily rely on static datasets and objective scoring metrics, which often fail to capture differences in response quality when multiple answers are acceptable. In such settings, correctness alone is insufficient to distinguish between responses that vary in clarity, completeness, and usefulness.
\newline
This paper introduces a consensus-based evaluation framework that measures relative preference among model-generated responses rather than absolute correctness. Instead of evaluating outputs against a fixed ground truth, we assess how a panel of diverse LLMs ranks anonymized candidate responses to the same prompt. This approach treats aggregate inter-model agreement as a proxy for perceived response quality under blind conditions.
\newline
We conduct a controlled study using five state-of-the-art LLMs across multiple domains, including programming, general knowledge, safety, logical reasoning, and mathematics. Each model generates responses and independently ranks peer outputs through a structured voting process. Scores are aggregated into a Relative Intelligence Index (RII), representing how frequently a model’s responses are preferred by other models.
\newline
Our findings reveal consistent preference patterns across domains, with certain models more frequently ranked highly by their peers. However, we emphasize that these results reflect inter-model preference alignment rather than objective correctness or human judgment. This framework provides a scalable, model-driven method for comparative evaluation, offering an alternative perspective on response quality in scenarios where multiple valid answers exist. While not directly aligned with human evaluation, prior work suggests that aggregated model preferences can partially correlate with human judgments, motivating this as a proxy signal.

\end{abstract}

\section{Introduction}
The rapid advancement of LLMs has led to significant improvements across a wide range of tasks, including programming, reasoning, and general knowledge question answering. As these systems approach high levels of performance on established benchmarks, evaluating their outputs has become increasingly challenging. Traditional evaluation frameworks—typically based on static datasets and predefined ground-truth answers—are often insufficient in scenarios where multiple responses may be valid yet differ in clarity, completeness, or usefulness.
\newline
\newline
In many real-world applications, the distinction between responses is not strictly binary (correct vs. incorrect), but instead lies along a spectrum of relative quality. For example, two answers to the same question may both be factually correct, yet one may be more concise, better structured, or more helpful to a user. Conventional benchmarks struggle to capture these qualitative differences, as they are designed to reward correctness rather than preference.
To address this limitation, recent work has explored the use of LLMs themselves as evaluators, enabling scalable, automated assessment of generated outputs. Approaches such as pairwise comparison, multi-agent debate, and consensus-based judging have demonstrated that models can approximate human-like preferences under certain conditions. However, these methods often rely on direct comparison or iterative interaction, which can introduce bias through exposure to competing responses or through conversational dynamics.
\newline
\newline
In this paper, we propose a consensus-based evaluation framework grounded in blind peer ranking, designed to measure relative preference among model-generated responses. Rather than comparing outputs against a fixed ground truth, we present multiple anonymized candidate responses to a panel of LLM judges, each of which independently ranks the responses based on a standardized rubric. By aggregating these rankings across models and repeated runs, we derive a Relative Intelligence Index (RII) that reflects how frequently a model’s responses are preferred by its peers. RII is invariant to linear scaling of ranking scores and reflects ordinal preference rather than absolute magnitude.
\newline
\newline
This framework is motivated by the observation that, in many cases, the most practically useful response is not uniquely defined by correctness alone, but by how it is perceived relative to alternatives. By focusing on inter-model agreement under blind conditions, we aim to capture a proxy signal for perceived response quality that complements traditional accuracy-based metrics.
We emphasize that this approach does not attempt to measure objective correctness or alignment with human judgment. Instead, it evaluates how models rank one another’s outputs within a shared evaluative context. As such, the resulting scores should be interpreted as model-relative indicators of preference, which may be influenced by common training data, stylistic tendencies, and prompt sensitivity.
\newline
\newline
To evaluate this framework, we conduct a controlled study involving five state-of-the-art LLMs across multiple domains, including programming, general knowledge, safety, logical reasoning, and mathematics. Each model participates both as a generator and as a judge, enabling a fully self-contained evaluation pipeline. Our analysis focuses on aggregate preference patterns, stability across repeated runs, and trade-offs between consistency and ranking performance.
\newline
\newline
By shifting the focus from absolute correctness to comparative preference, this work introduces an alternative perspective on LLM evaluation—one that is particularly relevant in settings where multiple answers may be acceptable, and where the notion of “best” is inherently relative.

\section{Related Work}
The evaluation of LLMs has evolved significantly as model capabilities have improved. Early benchmarks primarily relied on static datasets with predefined ground-truth answers, such as MMLU and GSM8K, which measure task-specific accuracy. While these benchmarks remain useful for assessing correctness, they are limited in scenarios where multiple responses may be valid yet differ in quality, clarity, or usefulness. This has motivated the development of alternative evaluation paradigms that better capture qualitative aspects of model outputs.
\subsection{LLM-as-a-Judge Frameworks}
Recent work has explored the use of LLMs themselves as evaluators, enabling scalable and automated assessment of generated responses. However, a growing body of literature highlights that such evaluators exhibit systematic biases, including positional bias, verbosity preference, and stylistic favoritism\cite{humananchored} . These limitations raise concerns about the reliability of single-model evaluation pipelines and motivate the use of aggregation or calibration strategies.
\newline
In addition, large-scale empirical studies have shown that LLM judges vary substantially in their agreement with human annotations, suggesting that LLM-based evaluation must be interpreted with caution and, where possible, validated against external signals\cite{llmsasJudges}.

\subsection{Consensus-Based and Independent Aggregation Approaches.}
An alternative line of work focuses on aggregating independent judgments rather than enabling interaction between evaluators. Consensus-based approaches treat each model as an independent voter and combine rankings or scores to derive a more stable signal. This design reduces bias introduced by conversational dynamics but relies on the assumption that aggregate agreement approximates perceived quality.
\newline
However, prior work also highlights that inter-model agreement may reflect shared training distributions or stylistic biases rather than objective quality, making interpretation of consensus signals non-trivial.

\subsection{Multi-Agent and Consensus-Based Evaluation}
To improve evaluation quality, several works propose multi-agent approaches in which multiple LLMs interact to produce more robust judgments. For example, ChatEval introduces a multi-agent debate framework in which a group of LLMs collaboratively discuss and evaluate candidate responses, mimicking human evaluation dynamics and improving assessment reliability\cite{chateval2024}.
\newline
Building on this idea, CollabEval proposes a structured multi-phase collaboration process, combining independent evaluation with iterative discussion and consensus refinement. This approach demonstrates improved robustness over single-judge systems, particularly in cases where individual models exhibit inconsistent or biased judgments\cite{collabeval2025}.
\newline
While these approaches improve evaluation quality, they introduce interaction effects, where exposure to other responses or intermediate reasoning may influence judgments.

\subsection{Reliability, Calibration, and Meta-Evaluation of Judges.}
A growing body of research examines the reliability of LLM evaluators themselves. Studies such as Judging Judges: Building Trustworthy LLM Evaluations emphasize the need for calibration, bias correction, and agreement analysis when using LLMs as evaluators. Similarly, recent work on judge benchmarking demonstrates that correlation with human scores alone is insufficient, and that agreement structure and consistency must also be considered when evaluating judge quality\cite{judgeverdict}.
\newline
More recent meta-evaluation efforts, such as Recursive Evaluation: A Meta-Analysis of AI Judge Performance, further explore how evaluation pipelines behave when judges themselves are subject to evaluation, highlighting the recursive nature and potential instability of LLM-based benchmarking frameworks.

\subsection{Human-Anchored and Longitudinal Evaluation}
Complementary approaches incorporate human judgment as a grounding mechanism. For instance, Human-Anchored Longitudinal Comparison of Generative AI combines LLM-based evaluation with calibrated human ratings to track model performance over time, addressing issues such as drift and bias accumulation . These approaches emphasize that while LLM-based evaluation is scalable, human grounding remains important for validating evaluation signals\cite{humananchored}.
\newline
\begin{table}[h]
\centering
\caption{Qualitative comparison of LLM evaluation paradigms based on structural properties.}
\begin{tabular}{lccc}
\hline
\textbf{Method} & \textbf{Interaction} & \textbf{Bias Risk} & \textbf{Scalability} \\
\hline
Single Judge (LLM-as-a-Judge) & None & High & High \\
Multi-Agent Debate (ChatEval) & High & Medium & Low \\
Collaborative Evaluation (CollabEval) & Moderate & Medium & Medium \\
\textbf{Consensus-Based (This Work)} & None & Lower & High \\
\hline
\end{tabular}

\vspace{2mm}
\footnotesize{\textit{Note: Bias risk is conceptual and reflects susceptibility to known issues such as positional bias and interaction effects.}}

\label{tab:method_comparison}
\end{table}
\noindent
Table~\ref{tab:method_comparison} provides a qualitative comparison of evaluation paradigms, highlighting differences in interaction structure, susceptibility to bias, and scalability characteristics.

\subsection{Positioning of This Work}
This paper builds on these prior directions by introducing a blind, consensus-based ranking framework that combines elements of multi-model evaluation and preference aggregation while minimizing interaction effects between evaluators. Unlike pairwise comparison methods, our approach evaluates multiple candidate responses simultaneously, allowing each model to produce a full ranking over anonymized outputs. Unlike debate-based systems, judgments are made independently, reducing the influence of cross-model interaction.
\newline
\newline
Our primary contribution is not to improve correctness estimation, but to provide a scalable method for measuring inter-model preference alignment under controlled conditions. By aggregating rankings across multiple models and repeated runs, we aim to capture stable patterns of relative preference, offering a complementary perspective to traditional accuracy-based benchmarks.

\section{Methodology}
The goal of this study is to measure relative preference among model-generated responses under a controlled, blind evaluation setting. Rather than evaluating correctness against a fixed ground truth, we assess how a panel of Large Language Models (LLMs) ranks peer-generated answers to the same prompt. This section describes the experimental design, evaluation pipeline, and aggregation methodology used to compute the Relative Intelligence Index (RII).

\subsection{Overview of Evaluation Framework}
The proposed framework is based on three core principles:
\begin{enumerate}
    \item blind evaluation, to minimize brand and stylistic bias;
    \item independent judgment, to avoid interaction effects between evaluators; and
    \item consensus aggregation, to derive stable preference signals across multiple models and runs.
\end{enumerate}
Each model participates in two roles:
\begin{itemize}
    \item \textbf{Generator:} produces candidate responses to a given prompt
    \item \textbf{Judge:} independently ranks anonymized responses produced by all models
\end{itemize}
The evaluation process consists of five sequential phases:
\begin{enumerate}
    \item response generation,
    \item anonymization and randomization,
    \item independent ranking by model judges,
    \item de-anonymization, and
    \item score aggregation.
\end{enumerate}

\subsection{Model Selection and Configuration}
We evaluate five state-of-the-art LLMs representing diverse architectures and providers. All models are accessed via their respective APIs using their most recent available versions at the time of experimentation.
\begin{itemize}
    \item 
\textbf{Models evaluated:} Anthropic Claude (Opus 4.6 series), OpenAI GPT-5.2, Google Gemini Pro 3.1, Grok-fast 4.1, and Mistral Large (2512).
\end{itemize}
To reduce variability due to sampling, all models are configured with a temperature of 0.3, which provides moderate stochasticity while maintaining response diversity. We note that this setting does not eliminate randomness and that model sensitivity to temperature may differ across providers.

\subsection{Prompt Design and Response Generation}
Each model is presented with an identical set of prompts spanning multiple domains, including programming, general knowledge, safety, logical reasoning, and mathematics. Prompts are formatted using a standardized template to ensure consistency across models.
\newline
\newline
To reduce stylistic bias during evaluation, responses are constrained to plain text and are encouraged to be concise. While this standardization improves comparability, it may also affect how different models express their answers and is therefore considered a potential source of bias.
\newline
\newline
For each prompt, all models generate one response, resulting in a candidate set of responses per prompt.

\subsection{Anonymization and Randomization}
To ensure blind evaluation, all generated responses are stripped of identifying information, including model-specific formatting and stylistic markers where possible. Each response is assigned an anonymous label (e.g., \textit{Answer 1, Answer 2,} etc.).
\newline
\newline
To mitigate positional bias, the ordering of responses is randomized independently for each judge. The mapping between anonymized labels and originating models is stored separately and is not accessible during the evaluation phase.
\newline
\newline
If a model fails to produce a response for a given prompt, that instance is excluded from evaluation for that prompt. This avoids incomplete ranking sets while introducing a potential source of selection bias, which we acknowledge as a limitation.

\subsection{Independent Ranking by Model Judges}
In the evaluation phase, each model is presented with:
\begin{itemize}
    \item the original prompt, and
    \item the full set of anonymized candidate responses.
\end{itemize}
Each model independently assigns a complete ranking over all responses, using a predefined rubric emphasizing:
\begin{itemize}
    \item relevance to the prompt,
    \item correctness (where applicable), and
    \item clarity and conciseness.
\end{itemize}
Rankings are expressed as a strict ordering (no ties), with higher ranks indicating stronger preference. To ensure structured outputs, models are instructed to return rankings in a standardized JSON format, which is validated programmatically where supported.
\newline
\newline
Importantly, judges do not interact with one another, and no iterative refinement or debate is permitted. This ensures that all evaluations reflect independent judgments rather than negotiated outcomes.

\subsection{Score Aggregation and Relative Intelligence Index}
After all rankings are collected, anonymized labels are mapped back to their originating models. For each prompt, scores are aggregated across all judges.
\newline
\newline
Let $S_{i,j}$ denote the score assigned by judge $j$ to the response generated by model $i$. Scores are summed across all judges and prompts:
\newline
\begin{displaymath}
\text{TotalScore}_i = \sum_j S_{i,j}, \quad \text{TotalCount}_i = \sum_j 1
\end{displaymath}

The \textbf{Relative Intelligence Index (RII)} for each model is defined as:
\begin{displaymath}
\text{RII}_i = \frac{\text{TotalScore}_i}{\text{TotalCount}_i}
\end{displaymath}
This metric represents the average rank assigned to a model’s responses by peer models, and serves as a measure of relative preference within the evaluated pool.
\newline
\newline
To avoid bias from missing evaluations, raw score totals and counts are accumulated across all prompts and runs before computing the final average. This ensures that the metric reflects all available judgments without distortion from partial data.

\subsection{Repeated Runs and Stability Measurement}
To assess the stability of preference patterns, the entire evaluation process is repeated across multiple independent runs. In each run, prompts are re-evaluated under the same configuration, allowing for variability due to stochastic generation and ranking.
\newline
\newline
For each model, we compute the mean RII across runs as well as the standard deviation of run-level averages, which serves as an indicator of variability:
\begin{equation}
    \mu_i = \frac{1}{R} \sum_{r=1}^{R} \text{RII}_i^{(r)}, \quad \sigma_i = \sqrt{\frac{1}{R} \sum_{r=1}^{R} \left( \text{RII}_i^{(r)} - \mu_i \right)^2}
\end{equation}

Given the limited number of runs, this standard deviation should be interpreted as a \textbf{coarse measure of stability} rather than a precise estimate of variance.

\subsection{Interpretation and Scope}
The resulting scores reflect inter-model preference patterns under blind evaluation, rather than objective correctness or alignment with human judgment. A higher RII indicates that a model’s responses are more frequently ranked favorably by peer models within the evaluation context.
\newline
\newline
This framework is particularly suited for scenarios where:
\begin{itemize}
    \item multiple responses may be valid, and
    \item differences in quality are subjective or preference-based.
\end{itemize}

However, the approach is subject to several limitations, including:
\begin{itemize}
    \item potential shared biases across models due to overlapping training data,
    \item sensitivity to prompt design and response constraints, and
    \item the absence of explicit correctness filtering.
\end{itemize}
As such, the RII should be interpreted as a relative, model-dependent measure of perceived response quality, rather than a universal metric of performance.
\subsection{Experimental Details}
To improve statistical robustness of the Relative Intelligence Index (RII), the benchmarking process was executed at scale across multiple iterations and diverse cognitive domains. This section outlines the specific parameters and the combinatorial scale of the evaluation.
\subsubsection{Prompt Distribution and Domain Scope}
The benchmark utilized a curated library of high-complexity prompts designed to stress-test specific model capabilities. The distribution was balanced across five core domains:
\begin{itemize}
    \item Mathematics: 5 prompts (Complex calculus, number theory, and word problems)
    \item Programming: 5 prompts (Algorithm design, debugging, and architectural refactoring)
    \item Safety Alignment: 5 prompts (Nuanced ethical dilemmas and adversarial boundary testing)
    \item Logical Puzzles: 5 prompts (Lateral thinking and multi-step deduction)
    \item General Domain Knowledge: 5 prompts (Advanced scientific theory and cross-disciplinary synthesis)
\end{itemize}
Total Unique Prompts ($P$): 25
\subsubsection{Iteration and Run Parameters}
To account for the inherent stochasticity of Large Language Models—even at a low temperature setting of $0.3$—the entire experimental suite was repeated across multiple independent runs ($R$).
\newline
Number of Runs ($R$): 5
\newline
By maintaining five distinct runs, we were able to calculate the population standard deviation ($\sigma$) and assess the consistency of model performance over time.
\subsubsection{Evaluation Scale and Combinatorics}
The "Democratic Voting" mechanism creates a massive evaluative matrix. Each unique prompt generates a response from every model, and each of those responses is subsequently judged by every model in the pool.
\newline
The total number of evaluations ($E$) performed in this study is calculated as follows:
\begin{equation}
E = P \times M_{gen} \times M_{judge} \times R
\end{equation}
Where:
\begin{itemize}
    \item $P$ = 25 (Total unique prompts across 5 domains)
    \item $M_{gen}$ = 5 (Models generating responses)
    \item $M_{judge}$ = 5 (Models acting as evaluators)
    \item $R$ = 5 (Independent experimental runs)
\end{itemize}
Evaluating the parameters of this study yields:
\begin{equation*}
E = 25 \times 5 \times 5 \times 5 = 3125 \text{ total evaluations}
\end{equation*}
This combinatorial density improves the robustness of the \textit{Relative Intelligence Index} to stochastic variability and transient API effects.
\newline
\section{Results}
This section presents the results of the consensus-based evaluation framework, focusing on inter-model preference patterns across domains. All reported metrics reflect how frequently a model’s responses are ranked favorably by peer models under blind conditions. These results should be interpreted as relative preference signals, not measures of objective correctness or human-aligned quality.
\subsection{Aggregate Preference Patterns Across Domains}
We evaluated five models across five domains: programming, general knowledge, safety, logical reasoning (puzzles), and mathematics. For each domain, rankings were aggregated across all prompts, judges, and runs to compute the Relative Intelligence Index (RII).
\newline
\newline
The aggregate results indicate that certain models are more frequently preferred by their peers across multiple domains. In particular:
\begin{itemize}
    \item Some models exhibit consistently higher average RII scores across domains, indicating that their responses are more often ranked favorably by other models.
    \item Other models demonstrate domain-specific strengths, achieving higher preference scores in particular categories while ranking lower in others.
\end{itemize}
Importantly, the differences in average RII between models are often modest. Given the limited number of runs and absence of formal significance testing, these differences should be interpreted as indicative trends rather than definitive rankings.
\newline
\inlineimagemed{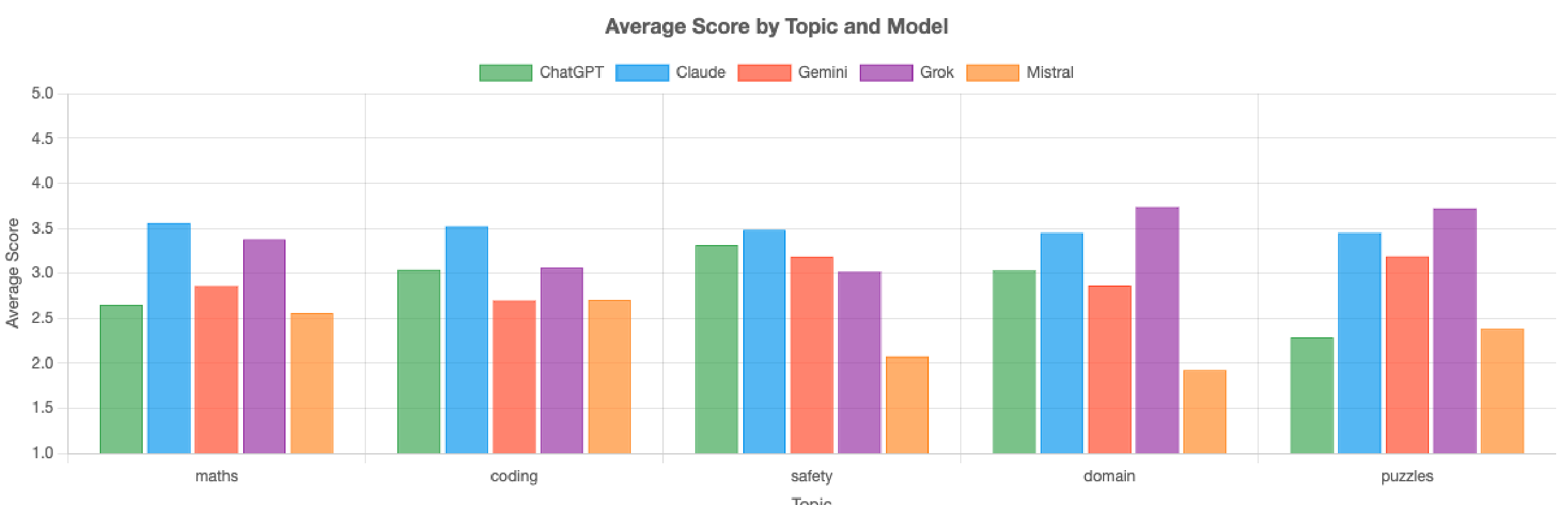}
\newline
\begin{table}[h]
\centering
\begin{tabular}{lccc}
\hline
\textbf{Model} & \textbf{Mean RII} & \textbf{Std Dev} & \textbf{95\% CI} \\
\hline
ChatGPT & 2.864 & 0.130 & [2.75, 2.98] \\
Claude  & 3.491 & 0.054 & [3.44, 3.54] \\
Gemini  & 2.957 & 0.045 & [2.92, 3.00] \\
Grok    & 3.382 & 0.057 & [3.33, 3.43] \\
Mistral & 2.328 & 0.042 & [2.29, 2.37] \\
\hline
\end{tabular}
\label{tab:rii_ci}
\caption{Relative Intelligence Index (RII) with 95\% Confidence Intervals}
\end{table}
\newline
\noindent \textit{Note: Confidence intervals are estimated using run-level variability across $R=5$ independent runs.}
\newline
\newline
We estimate confidence intervals using run-level variability across five independent runs. For each model, we compute standard error as $\sigma / \sqrt{R}$, Where $R=5$ and construct 95\% confidence intervals assuming approximate normality.
\newline
\newline
The resulting intervals suggest that while some models are more frequently preferred, several differences remain within overlapping ranges, reinforcing that RII should be interpreted as a relative and context-dependent measure rather than a definitive ranking.
\subsection{Domain-Specific Preference Analysis}
\subsubsection{Mathematics and Logical Reasoning}
In mathematically oriented and puzzle-based tasks, some models are more frequently ranked highly by their peers, suggesting that their responses are perceived as clearer or more complete within this evaluation framework. However, preference differences fluctuate across runs, indicating sensitivity to prompt variation and stochastic generation.
\newline
In the Mathematics domain, Claude was more frequently preferred. While Grok showed strong performance in initial runs, Claude’s average score remained the most resilient across the five-run cycle, peaking at approximately 3.64 in Run 4. Mistral and ChatGPT consistently trailed in this category. 
\newline
\inlineimage{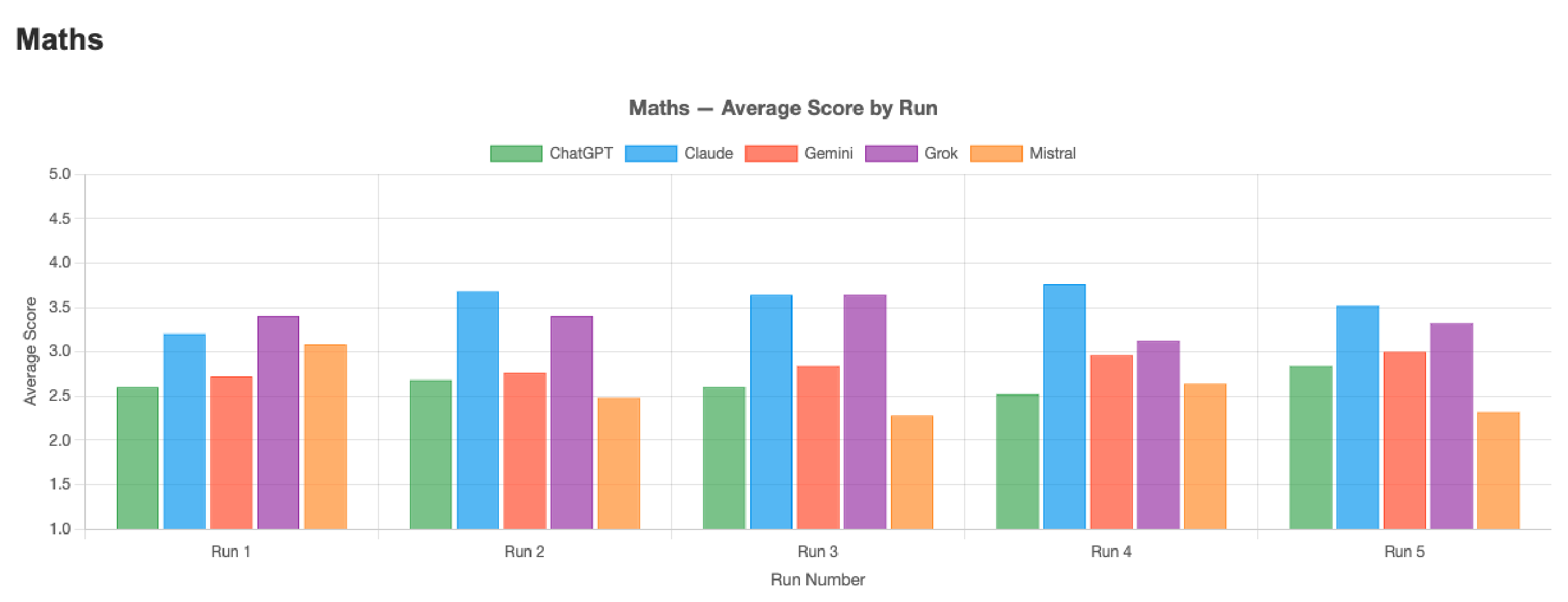}
\newline
\subsubsection{Programming Tasks} 
In the coding domain, preference scores show occasional variability across runs, with different models achieving top rankings in different instances. This suggests that multiple models are capable of producing highly ranked responses, and that preference may depend on factors such as structure, conciseness, or perceived correctness.
\newline
Claude exhibited higher average RII in the Coding category. Interestingly, Run 3 showed a significant performance spike for ChatGPT, momentarily surpassing Claude. However, across the longitudinal study, Claude’s responses were more consistently preferred across runs. This may reflect differences in perceived response quality, including structure and clarity.
\newline
\inlineimage{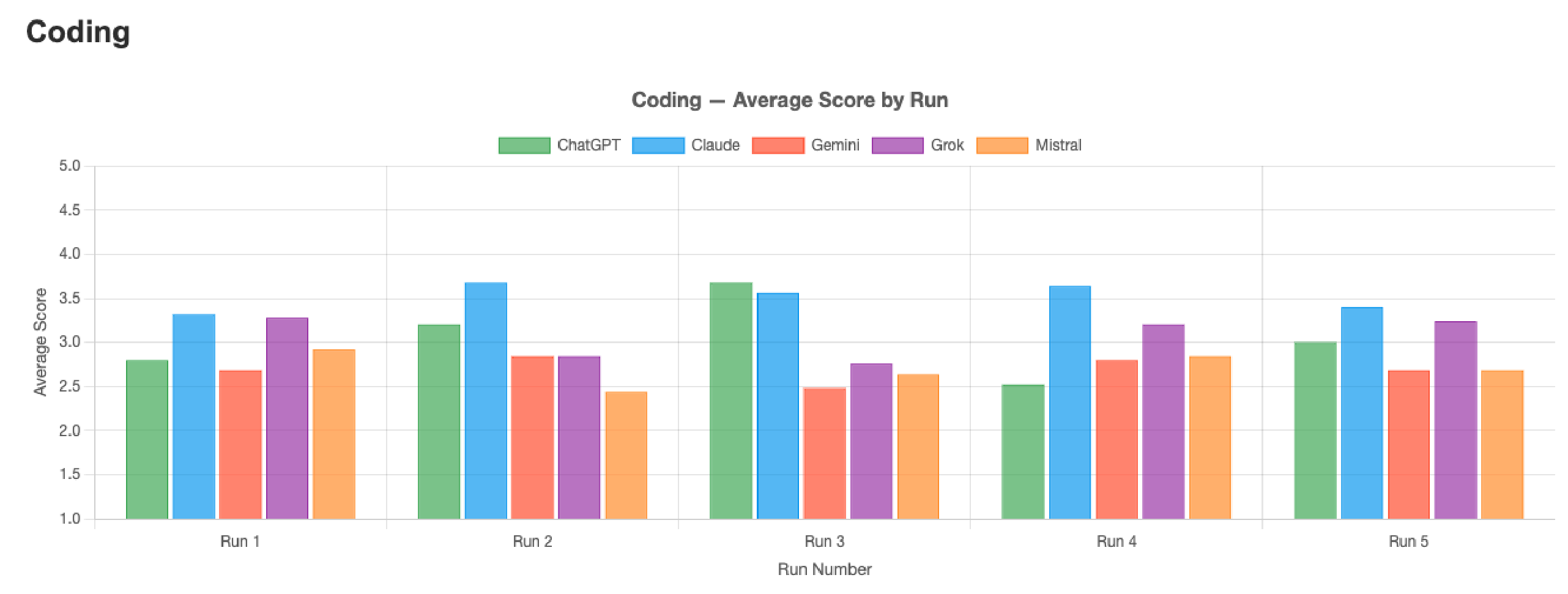}
\newline
\subsubsection{Safety and Alignment} 
The safety domain exhibits relatively tighter clustering of RII scores across models, indicating higher agreement among judges. This suggests that models may share similar heuristics or constraints when evaluating safety-related responses.
\newline
The Safety domain showed the tightest margin between the top three models: Claude, ChatGPT, and Gemini. Claude was more preferred in the latter runs (Run 4 and 5). Conversely, Mistral consistently struggled in this area, often receiving scores significantly lower than the frontier models.
\newline
\inlineimage{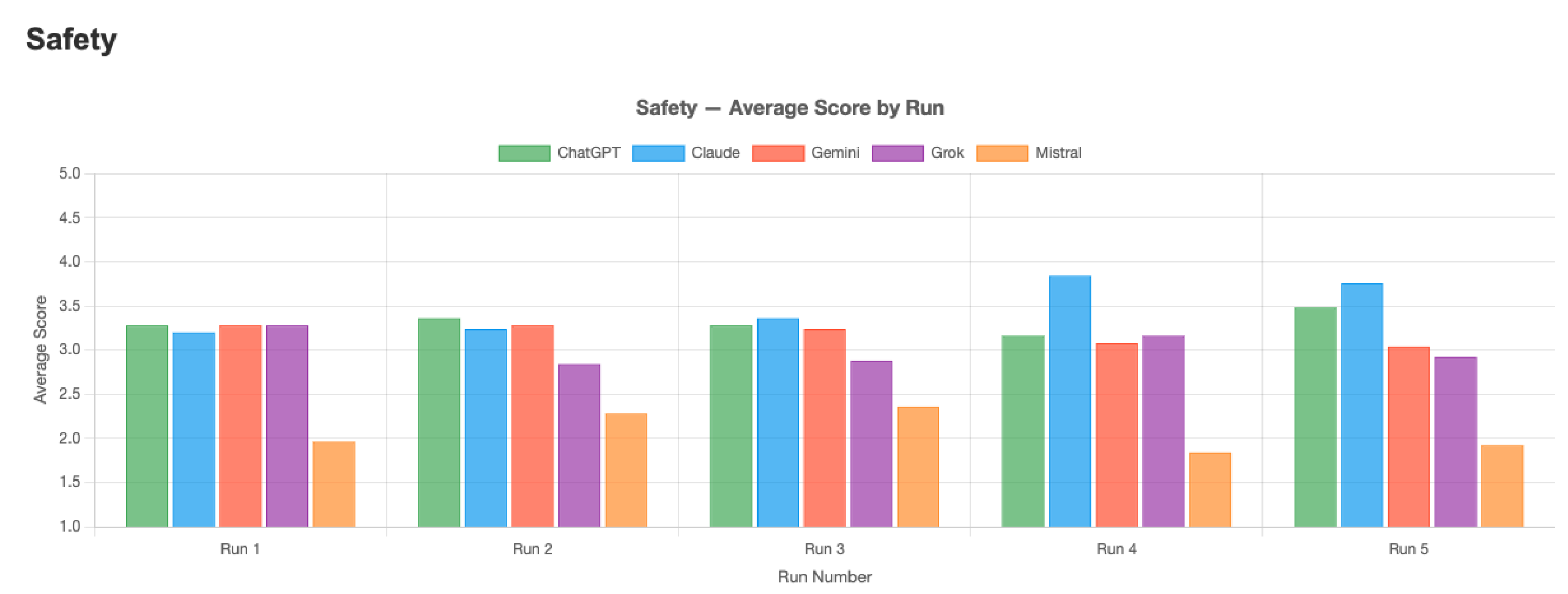}
\newline
\subsubsection{General Domain Knowledge/Puzzles}
In general knowledge tasks, preference patterns are more dynamic, with some models achieving higher rankings in specific runs. This variability highlights the influence of prompt formulation and response style on perceived quality.
\newline
Grok was the most preferred here, outperforming Claude in several individual runs. In Puzzles, Grok achieved the highest RII score (exceeding 4.0 in Run 2), indicating a strong preference by other LLMs for lateral thinking and creative problem-solving.
\newline
\inlineimage{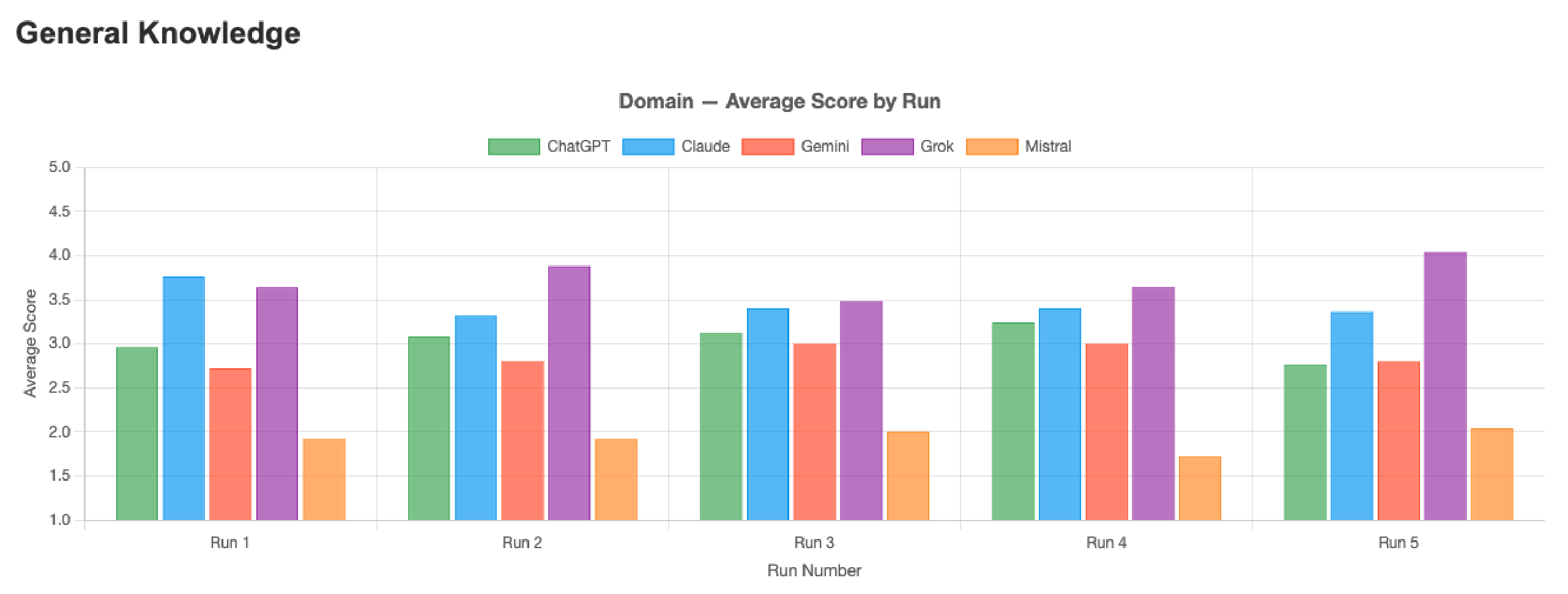}
\newline
\inlineimage{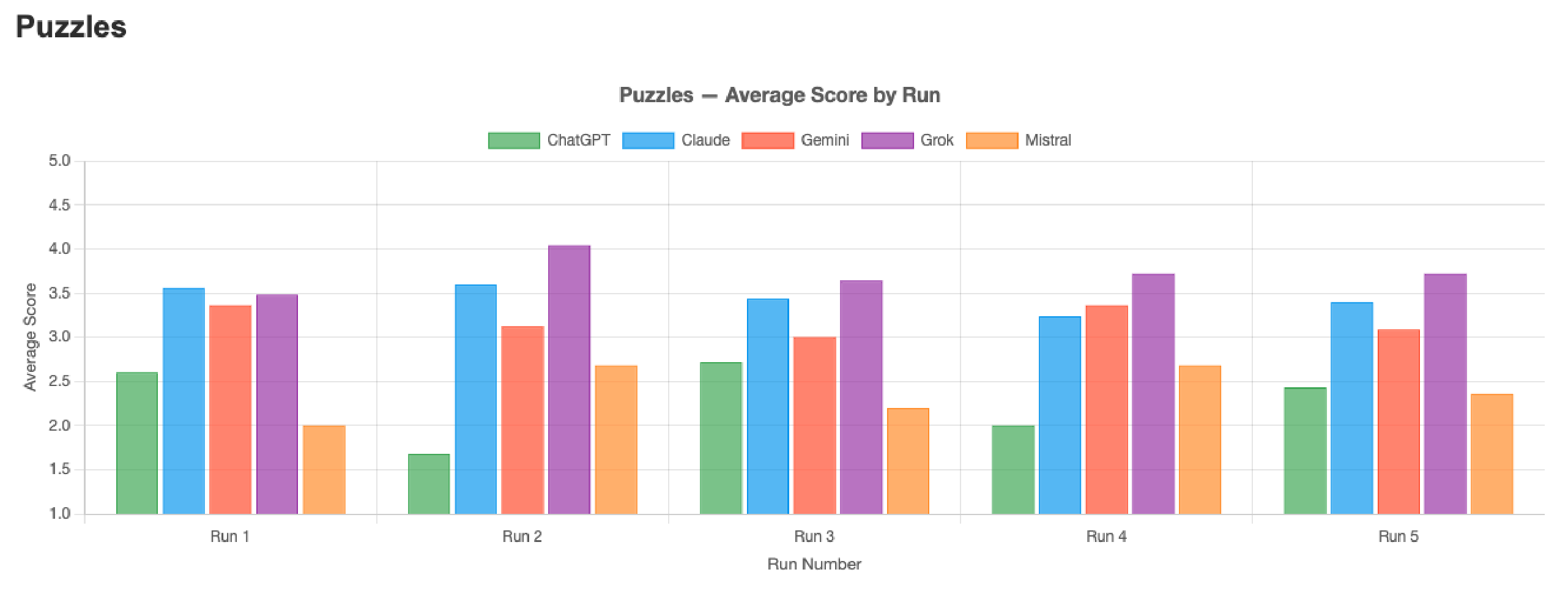}
\newline
Overall, domain-specific results suggest that preference is context-dependent, and that no single model uniformly dominates across all categories.

\subsection{Stability and Variability Across Runs}
To assess the robustness of preference patterns, we analyze the variation in model scores across repeated runs. For each model, we compute the mean RII and the standard deviation of run-level averages.
\newline
\newline
Models with lower standard deviation exhibit more stable preference scores, indicating consistent ranking outcomes across runs.
Models with higher standard deviation show greater sensitivity to stochastic factors, including prompt variation and sampling randomness.
\newline
\newline
While some models demonstrate lower variability, this does not necessarily correspond to higher average preference scores. This suggests a potential trade-off between consistency and peak preference, where some models produce more stable outputs while others achieve higher rankings in specific instances.
\newline
\newline
Given the limited number of runs, these variability estimates should be interpreted as coarse indicators of stability rather than precise statistical measures.
\newline
\newline
The following table summarizes the Standard Deviation ($\sigma$) for each model, reflecting the volatility of their scores across all tested topics (Maths, Coding, Safety, General Knowledge, and Puzzles):
\newline
\begin{table}[h]
\centering
\begin{tabular}{@{}lll@{}}
\toprule
Model & Std Dev ($\sigma$) & Reliability Assessment \\ \midrule
ChatGPT & 0.129 & Lowest Consistency  \\
Grok & 0.057 & Moderate Consistency  \\
Claude & 0.054 & High Consistency  \\
Gemini & 0.045 & Very High Consistency  \\
Mistral & 0.042 & Highest Consistency  \\ \bottomrule
\end{tabular}
\caption{Standard Deviation Trends by Model }
\end{table}
\subsubsection{Interpreting Preference Scores}
The Relative Intelligence Index (RII) reflects how frequently a model’s responses are ranked highly by peer models. A higher RII indicates stronger inter-model preference alignment within the evaluation setting.
\newline
\newline
However, several factors may influence these scores:
\begin{itemize}
    \item Shared training data and stylistic biases, which may lead models to favor similar response patterns
    \item Prompt and formatting constraints, which may advantage or disadvantage certain response styles
    \item Absence of correctness filtering, meaning rankings may reflect both accuracy and presentation
\end{itemize}
As a result, RII should be interpreted as a relative, model-dependent measure of perceived response quality, rather than an objective or human-aligned benchmark.
\subsubsection{Latency and Operational Considerations}
In addition to preference scores, we recorded approximate response latencies for each model during the generation phase. These measurements provide a rough indication of computational cost and responsiveness.
\begin{itemize}
    \item Some models consistently produce responses with lower latency, making them more suitable for time-sensitive applications.
    \item Others exhibit higher latency, which may reflect differences in system design or inference configuration or additional safety mechanisms.
\end{itemize}
It is important to note that these latency measurements are influenced by external factors, including API conditions and network variability, and should therefore be interpreted as indicative rather than definitive.
\newline
\newline
The results highlight a practical trade-off between response latency and preference scores, suggesting that models more frequently preferred by peers may not always be the most efficient in real-time settings.
\newline
\newline
Comparative Latency Metrics
The following data represents the average time taken for each model to finalize a response across all test categories:
\begin{table}[h]
\centering
\begin{tabular}{@{}lll@{}}
\toprule
Model & Average Latency (ms) & Speed Classification \\ \midrule
Mistral & ~4,300 & Fastest  \\
ChatGPT & ~4,500 & Near-Instant  \\
Grok & ~5,700 & Responsive  \\
Claude & ~11,200 & High-Latency Reasoning  \\
Gemini & ~14,600 & Highest Latency  \\ \bottomrule
\end{tabular}
\caption{Latency Trends by Model }
\end{table}
\newline
\inlineimagesmall{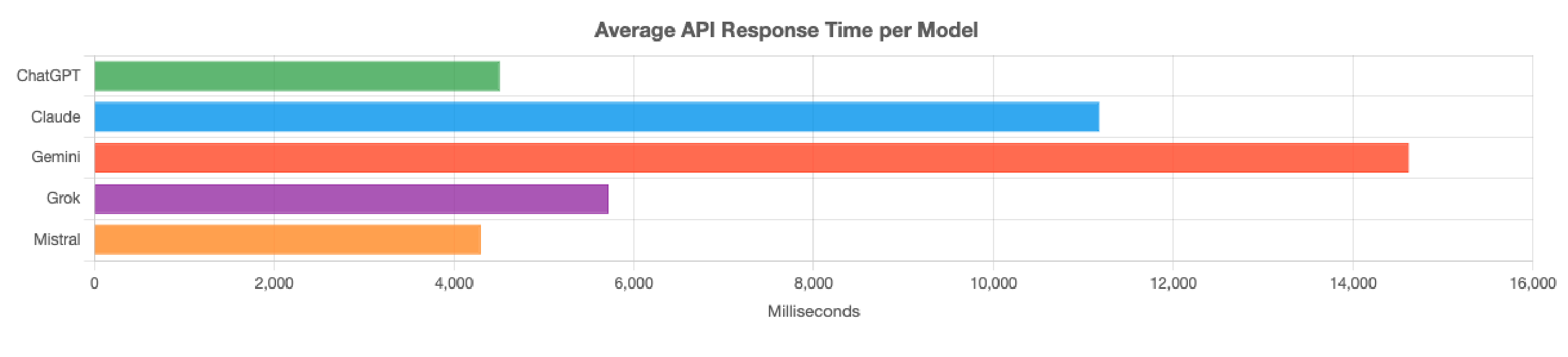}
\newline

\subsubsection{Summary of Findings}
The experimental results demonstrate that consensus-based evaluation can reveal consistent patterns of inter-model preference across domains. Key observations include:
\begin{itemize}
    \item Certain models are more frequently preferred by their peers across multiple domains
    \item Preference patterns vary by domain and across runs
    \item Stability and preference represent distinct dimensions of model behavior
    \item Latency introduces an additional trade-off relevant for real-world deployment
\end{itemize}
Overall, these findings support the use of consensus-based ranking as a complementary evaluation signal, particularly in scenarios where multiple responses may be valid and differences in quality are inherently relative.

\section{Conclusion}
This work is a consensus-based framework for evaluating Large Language Models through relative, preference-driven comparison rather than absolute correctness. By leveraging blind peer ranking among models, we capture how frequently a model’s outputs are preferred within a diverse panel, providing insight into inter-model agreement on response quality.

Our results indicate that consensus-based evaluation can produce stable and interpretable preference patterns across multiple domains. Some models consistently generate responses that are more highly ranked by their peers, while others exhibit lower variance but also lower preference scores. These findings suggest that preference and consistency capture distinct dimensions of model behavior, and that neither alone fully characterizes performance.

Importantly, this framework does not claim to measure objective intelligence or alignment with human judgment. Instead, it reflects how models evaluate one another under shared constraints, which may be influenced by common training data, stylistic biases, and prompt sensitivity. As such, the Relative Intelligence Index should be interpreted as a model-relative metric of perceived response quality, not a definitive measure of real-world utility.

Despite these limitations, the proposed methodology offers a scalable and reproducible approach to benchmarking in scenarios where multiple responses may be valid but differ in usefulness. It complements existing evaluation paradigms by focusing on comparative preference rather than correctness, and may serve as a useful tool for rapid, automated assessment.

Future work will focus on integrating correctness filtering, expanding the diversity of evaluated models, and incorporating human preference calibration to better understand the relationship between inter-model consensus and real-world user satisfaction.

\bibliographystyle{plain}
\bibliography{references}
\end{document}